# Time Series Classification: Lessons Learned in the (Literal) Field while Studying Chicken Behavior

Alireza Abdoli [1], Amy C. Murillo [2], Alec C. Gerry [2], Eamonn J. Keogh [1]
Department of Computer Science and Engineering, Department of Entomology
University of California, Riverside
Riverside, CA USA
aabdo002@ucr.edu, amy.murillo@ucr.edu, alecg@ucr.edu, eamonn@cs.ucr.edu

*Abstract*— Poultry farms are a major contributor to the human food chain. However, around the world, there have been growing concerns about the quality of life for the livestock in poultry farms; and increasingly vocal demands for improved standards of animal welfare. Recent advances in sensing technologies and machine learning allow the possibility of monitoring birds, and employing the lessons learned to improve the welfare for all birds. This task superficially appears to be easy, yet, studying behavioral patterns involves collecting enormous amounts of data, justifying the term *Big Data*. Before the big data can be used for analytical purposes to tease out meaningful, well-conserved behavioral patterns, the collected data needs to be pre-processed. The pre-processing refers to processes for cleansing and preparing data so that it is in the format ready to be analyzed by downstream algorithms, such as classification and clustering algorithms. However, as we shall demonstrate, efficient pre-processing of chicken big data is both non-trivial and crucial towards success of further analytics.

*Keywords*— *Time Series, Classification, Big Data, Machine Learning, Poultry Welfare.*

## I. Introduction

There was a time, not that long ago, when wearable devices were only the province of humans. How far we've come since then. There are high-tech wearables available, regardless of whether you're a tiny insect [1], grasshopper or a significantly larger animal like a cow [2]. In between come the chickens which are a major source of high-protein and low-fat food [3].

In recent years, there have been unprecedented technological advances in sensor technology, and sensors have become more affordable than ever. Consequently, sensor-driven data collection is increasingly becoming an attractive and practical option for researchers around the globe. Such collected sensor data is typically extracted in the form of time series data; which can be investigated with data mining techniques to summarize the behaviors of the animals [4].

Time series data is widely used in many domains and is of significant interest in data mining. In recent years researchers have proposed various algorithms for the efficient processing of time series datasets [5][6][7][8]. In this study, we use on-animal sensors to quantify specific behaviors performed by chickens. These behaviors, e.g. preening and dustbathing, are known or suspected to correlate with animal well-being.

## II. How The Data Grow Big?

Big Data refers to datasets which are big in size and volume; consequently, such datasets are challenging to process and analyze with traditional techniques and software in a timely manner. Datasets are growing rapidly because of the advances in the sensor technologies which enables cheap and mass collection of data. Big data size might range from a few terabytes to zettabytes of data [9]. Big Data applications encompass different fields from computer networks [10] and semantic web [11] to smart healthcare/clinical devices [12][13] and retail industry [14]. However, in order to interpret and analyze the data it should be prepared and cleansed, so that the data can be used by downstream algorithms such as clustering and classification algorithms. In this study, we discuss data management techniques for on-animal sensors to quantify specific behaviors performed by chickens.

## III. Big Data In Poultry Science

Given the ever-increasing population of the world, the demand for such food sources has been steadily growing. According to Food and Agriculture Organization of the United Nations (FAO), Poultry is the world's primary source of animal protein. Between 2000 and 2030, per capita demand for poultry meat is projected to increase by 271 percent in South Asia, 116 percent in Eastern Europe and Central Asia, 97 percent in the Middle East and North Africa and 91 percent in East Asia and the Pacific [3]. In developed countries, there are growing concerns about the ethical treatment of these animals; among which are housing conditions and how the animals are managed and treated.

Arthropod ectoparasites reside on the surface of the body of chickens, causing stress to the host, and potentially spreading to nearby chickens or other animal hosts [15]. Many of these ectoparasites, such as the northern fowl mite, adversely affect productivity (e.g. laying eggs) and health of the chickens. They may also impact poultry behavior and welfare. Understanding how the chicken behave (i.e. timing and frequency of chicken behaviors) can help producers determine behavior abnormalities due to infestations and deploy preventive and corrective control methods. We are not proposing that all chickens be monitored, that is clearly unfeasible. Our system is designed as a tool to allow researchers to assess the effects of various conditions on chicken health, and then use the lessons learned on the entire brood [4].

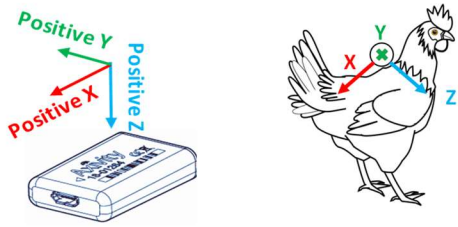

**Figure 1:** (*left*) Axivity AX3 axis alignment (*right*) positioning of AX3 sensor on the back of chicken.

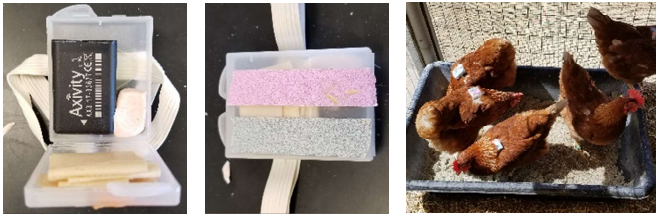

**Figure 2:** (*left*) Axivity AX3 sensor secured inside a plastic backpack with a rubber band wrist (*center*) The backpack ready to be mounted on a chicken with color markers for facilitated visual recognition (*right*) Chickens wearing backpacks on their back. Photos courtesy of A.C.M.

## IV. COLLECTING CHICKEN DATA

All chickens were housed and cared for in accordance with UC Riverside Institutional Animal Care and Use Protocol. Data is collected from chickens by placing the sensor on bird's back. The sensor is placed on back of the bird to allow for high-quality recording of various types of typical chicken behaviors, with the minimum interference and discomfort. The Axivity AX3 sensor used in our study, weighs about 11 grams and is configured with 100 Hz sampling frequency and +/- 8g sensitivity which allows for two weeks of continuous data collection with the battery fully charged. Figure 1 (*left*) shows the orientation of Axivity AX3 sensor (*right*) shows placement of sensor on the back of the chicken.

Figure 2 (*left*) shows the Axivity AX3 sensor secured inside a plastic backpack with a rubber band to allow placing the backpacks on the back of chickens (*center*) The backpack ready to mounted on the chicken with color markers for facilitated human recognition (*right*) shows placement of sensors on the back of the chickens. We used USB hubs for mass charging of the sensors; which allowed for simultaneous charging of multiple sensors towards saving human time and effort. After the sensors were fully charged the team member A.C.M connected each sensor to a desktop computer and utilized the AX3 GUI software to setup the sensors to start collecting data at some predefined date and time.

The overview of the processes to study chicken behavior and welfare is shown in Figure 3. This workflow repeated for each of the readings. The description of the individual processes is provided in the following:

1. The entomologist experts setup the sensors and make sure the sensors are fully charged and then the backpacks are mounted on the chickens.

2. The sensors collect data for 7-9 full days per reading. The data is stored in the CSV file format on the sensor's internal memory.

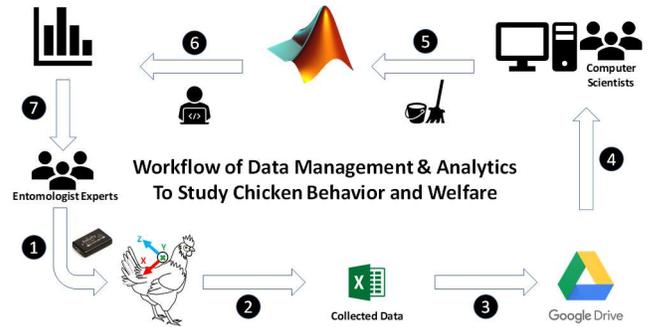

**Figure 3:** The workflow to study chicken behavior and welfare.

3. The entomologist experts pickup the sensors from chickens and upload the chicken data file (CSV file format) onto the Google Drive for facilitated accessibility and availability of data for all team members.

4. Following successful upload of data onto the Google Drive, computer scientists download the data.

5. Next, computer scientists proceed with the developed algorithm for cleansing and preparation of chicken data for further downstream processing (e.g. classification, clustering and etc.).

6. Given the cleansed and well-formatted data, the computer scientists can proceed with classification of the chicken behaviors. The classification results demonstrate the timing and frequency of various chicken behaviors throughout the 24-hours.

7. The classification results (i.e. behavior counts and frequency) are provided to entomologist experts so they can look for anomalies and irregularities in the chicken behavior and welfare.

## V. ALGORITHM FOR MANAGING CHICKEN BIG DATA

This section presents the algorithm developed for pre-processing and cleansing of data collected from sensors mounted on chickens. The chicken data downloaded off the sensors were into the format of single large CSV files per sensor holding data for the entire reading period (i.e. 7-9 days). The files were as big as 3-4 GB per file (several terabytes of chicken data); the large file sizes made them challenging to work with and process in MATLAB. Also, the entomologist experts preferred to work on individual days basis. So, an algorithm was developed to take in each CSV file and break it down into individual days. Also, the CSV file format was pretty space-consuming, so the developed algorithm stored the individual days as MATLAB friendly MAT files.

The most challenging part was to read the CSV files into memory and slice it into individual days; however, as the CSV files were large in volume, they would cause problems. In order to overcome the large size of CSV files we utilized the *Datastore* concept in MATLAB. The datastore essentially is a repository for collections of data that are too large to fit in memory. A datastore allows you to read and process data stored in multiple files on a disk, a remote location, or a database as a single entity. If the data is too large to fit in memory, you can manage incremental import of data for further processing [9].

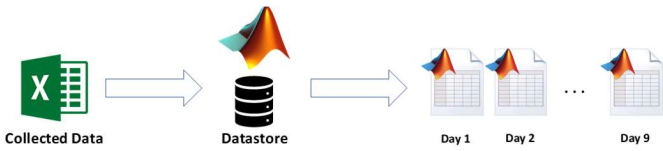

**Figure 4:** The slicer algorithm leverages the datastore in MATLAB and breaks the CSV file into individual days.

## VI. CHICKEN BEHAVIOR CLASSIFICATION

Given the data being cleansed and prepared, the computer scientist experts may proceed with classification of chicken behaviors for each full day. The computer scientists then provide the behavior counts to the entomologist experts so they can comment about potential anomalies in the behavior count and intervene with preventive and corrective actions.

The classification of chicken behaviors throughout the day has been discussed in [4]; in which computer scientists extracted a dictionary of well-preserved chicken behaviors (i.e. pecking, preening and dustbathing) from a short video annotated chicken dataset, as shown in Figure 5. Then, they used the dictionary of behaviors on the unseen datasets from other birds.

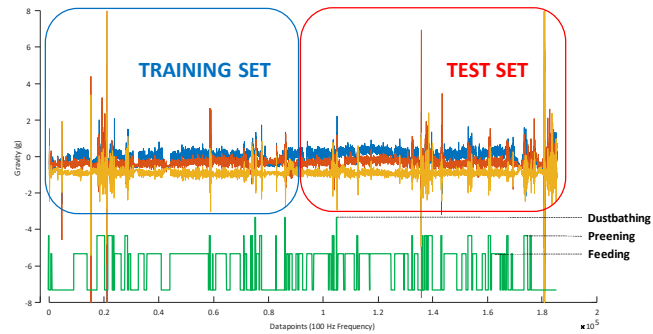

**Figure 5:** Three-dimensional chicken time series (the top/blue time series is X-axis; middle/red time series is Y-axis and bottom/yellow time series is Z-axis time series). The green lines represent annotations of observed chicken behaviors captured on video.

Following the classification of bird behaviors for every individual day the count and timing of behaviors are reported by the classification algorithm. This helps the entomologist experts to analyze the behavior count and timing and proceed with potential actions. Figure 6 shows the circadian plot on timing and frequency of a full-day chicken dataset.

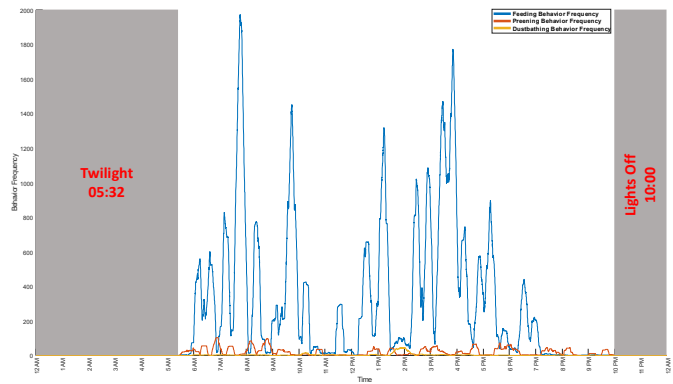

**Figure 6:** Temporal frequency of chicken behaviors (feeding/pecking, preening and dustbating) for the twenty-four hour chicken dataset.

## VII. CONCLUSIONS

In this paper we discussed processes involved with big data management. The suggested processes can be applied to any big dataset; while, this study applied the cleansing and pre-processing algorithm to the chicken datasets to classify and quantify chicken behaviors towards studying chicken welfare. Further, the paper briefly addressed classification and visualization of chicken behaviors such that the results are more intuitive to the audience.

## ACKNOWLEDGEMENT

We would like to acknowledge funding from "NSF 1510741–RI: Medium: Machine Learning for Agricultural and Medical Entomology (07/01/17-09/30/19)" and "USDA NIFA2017-67012-26100".